\documentclass[conference,compsoc]{IEEEtran}
\IEEEoverridecommandlockouts
\usepackage{cite}
\usepackage{amsmath,amssymb,amsfonts}
\usepackage{algorithmic}
\usepackage{graphicx}
\usepackage{textcomp}
\usepackage{xcolor}
\usepackage{url}
\usepackage{booktabs}
\usepackage{makecell}
\usepackage{pifont}
\def\BibTeX{{\rm B\kern-.05em{\sc i\kern-.025em b}\kern-.08em
    T\kern-.1667em\lower.7ex\hbox{E}\kern-.125emX}}

\begin{document}

\title{Beyond Capability Benchmarks: Learning Operational Fingerprints of LLM Cloud Services from Production Incident Metadata}

\author{
\IEEEauthorblockN{Meiwei Zhang, Eduardo Miranda, Bruce Baynes, Suvigya Jain, Wanlong Chen, Tao He, Sergey Borodavkin\textsuperscript{*}}
\IEEEauthorblockA{\textit{Google Cloud Platform, Google LLC} \\
Ireland, Canada, Switzerland \\
\textsuperscript{*}Corresponding author: sborodavkin@google.com}
}

\maketitle

\begin{abstract}
Managed LLM services are now part of real production systems, but model selection and service planning still rely heavily on capability benchmarks that reveal little about operational behavior after deployment. We present Operational Embedding (OpEmbed), a framework for learning compact operational fingerprints of LLM cloud services from structured, privacy-preserving support-case metadata, without using case text. OpEmbed aggregates model--time windows into an eight-channel operational signature and learns a low-dimensional representation via temporal contrastive learning, cross-view reconstruction, and generational-ordinality regularization. Evaluated on more than 33,000 production support cases spanning seven LLM families over 26 months at Google Cloud, OpEmbed recovers interpretable family- and version-level structure, improves leave-one-model-out operational forecasting over non-learned baselines, remains useful under limited early-window data, and supports cross-model fault-type transfer. We report the practical lessons learned from building and evaluating this tool for model onboarding, support readiness assessment, and operational monitoring.
\end{abstract}

\begin{figure}[htbp]
  \centering
  \fbox{
    \parbox{0.95\columnwidth}{
      \vspace{0.5em} 
      \textbf{Author's Note:} This paper was peer-reviewed and accepted by the IEEE International Symposium on Software Reliability Engineering (ISSRE) 2026 Industry Track. It was subsequently withdrawn from the official conference proceedings strictly due to corporate travel policy constraints preventing the required in-person presentation. We provide the full peer reviews and acceptance notification in the Appendix to certify its peer-reviewed status.
      \vspace{0.5em} 
    }
  }
\end{figure}


\section{Introduction}

Large language models (LLMs) are increasingly deployed as managed cloud services in enterprise environments \cite{manowska2026application,bokkena2024optimizing}. Operating such a service reliably is not only a question of model capability, but also of how the service behaves once it is live: which models generate more support load, what kinds of failures recur, how quickly issues are resolved, and how these patterns shift across versions. These are exactly the questions that platform and reliability teams have to answer for model onboarding, incident triage, support staffing, and launch readiness --- and they are questions that a capability leaderboard cannot answer.

A concrete scenario illustrates the gap. Suppose a platform team must decide which of two newly released models to promote as the default for a customer-facing product. Both score within a point of each other on public reasoning and coding benchmarks, so capability evaluation gives no basis for the decision. Three months after launch, one model turns out to generate roughly twice the support case volume of the other, with a disproportionate share escalated to engineering rather than resolved by first-line support. Nothing in a benchmark leaderboard would have anticipated this, yet it directly drives staffing needs, onboarding risk, and customer experience. This is the operational reliability gap that motivates our work: we want to anticipate and monitor this kind of divergence using signals that are already collected as a byproduct of running the service.

\begin{figure}[htbp]
    \centering
    \includegraphics[width=0.48\textwidth]{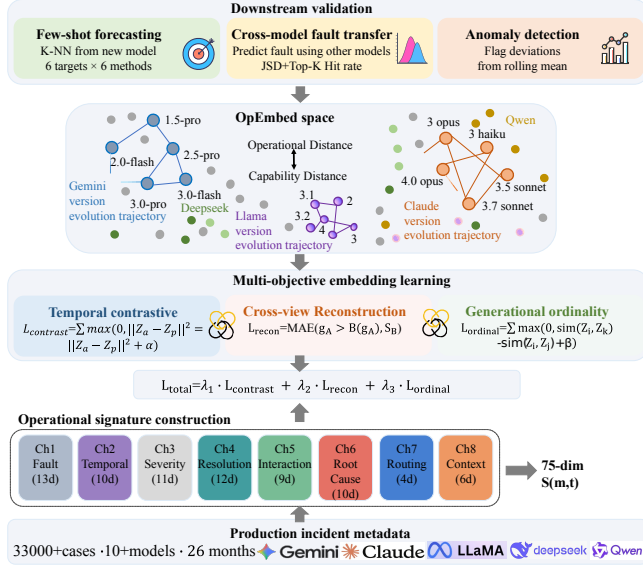}
    \caption{OpEmbed space framework.}
    \label{fig:image1}
\end{figure}

We argue that structured support-case metadata --- priority, fault category, timestamps, escalation patterns, routing behavior, and response-time statistics --- already describe a model's operational profile \cite{jiang2026representation}, without requiring access to case text or customer content \cite{rahman2022data}. Building on this observation, we present Operational Embedding (OpEmbed), shown in Fig.~\ref{fig:image1}, a framework that learns compact operational fingerprints of LLM cloud services from aggregated model--time windows.

Our contributions are: (1) we formulate LLM operational profiling as a representation-learning problem over structured incident metadata, defining eight signal channels spanning 75 features; (2) we combine temporal contrastive learning, cross-view reconstruction, and generational-ordinality constraints into a single training objective, and report results for this combined objective in our production-data evaluation, leaving a controlled decomposition of each term to follow-up work; (3) we evaluate the framework on more than 33,000 production support cases spanning seven LLM families over 26 months; and (4) we show that the learned representation supports leave-one-model-out forecasting, limited-history prediction, and cross-model fault transfer, and that operational similarity does not always follow vendor boundaries. From a reliability-engineering perspective, this provides a practical, tool-supported alternative to ad hoc vendor- or family-based heuristics for onboarding and support readiness planning.

\section{Related Work}

LLM evaluation is dominated by capability benchmarks such as MMLU \cite{hendrycks2021mmlu}, HumanEval \cite{chen2021codex}, HELM \cite{bommasani2023helm}, and more dynamic successors such as MMLU-Pro \cite{wang2024mmlupro}, Chatbot Arena \cite{chiang2024chatbot}, and LiveBench \cite{white2024livebench}. Surveys of this literature note that reproducibility, robustness, and real-world validity remain open challenges \cite{chang2024survey_eval,laskar2024systematic}, but these benchmarks primarily quantify capability, not how a model behaves operationally once deployed \cite{xia2024incorporating,wang2026llm}.

AIOps research offers the closest methodological context: alert handling, incident analysis, anomaly detection, and LLM-assisted root-cause diagnosis \cite{yu2024alerts_incidents,zhang2025aiops_llm,gupta2023logs_aiops,guo2024logformer,chen2024rcacopilot,roy2024rca_agents}. These methods typically diagnose individual incidents using logs or traces, and customer-support research typically classifies or routes individual tickets \cite{marcuzzo2022hierarchical_tickets,zangari2023ticket,zicari2021discovering}. In both cases, the unit of analysis is the individual case. OpEmbed instead operates at the model--time-window level, using structured metadata only, to learn a fingerprint that supports cross-model comparison, few-shot forecasting, and fault-pattern transfer --- capabilities that case-level tools do not target.

\section{Method}

\subsection{Problem formulation and intuition}

Our goal is a fixed-length summary of how a model behaves in production, at the granularity of a model and a time window, that is comparable across models and stable over time. Let $\mathcal{M}$ be deployed LLM models and $\mathcal{T}$ discrete time windows (calendar months in our deployment). For each model $m$ and window $t$ we observe support cases as structured records --- priority, fault category, timestamps, resolution outcome --- but never case text. We aggregate these into a signature $\mathbf{S}_{m,t}$ (Section~\ref{sec:signature}) and learn $f_\theta\colon \mathbb{R}^D \to \mathbb{R}^d$ producing an embedding $\mathbf{z}_{m,t}=f_\theta(\mathbf{S}_{m,t})$. Intuitively, $\mathbf{z}_{m,t}$ is a coordinate in an ``operational behavior space'': two models with nearby coordinates are expected to impose similar support burden, even across vendors, while a single model's coordinate should drift smoothly unless its deployment changes materially. We train $f_\theta$ to satisfy temporal smoothness, model discriminability, and generational ordinality.

\subsection{Operational signature construction}
\label{sec:signature}

For each model-month pair, we construct a 75-dimensional operational signature across eight channels: fault spectrum, temporal pattern, severity and customer mix, resolution efficiency, interaction dynamics, root cause, defect routing, and workload context. These channels were selected to match structured fields commonly populated in enterprise support systems, making the construction reusable in other ecosystems with comparable schemas. For sparse fields, such as root cause, we encode both observed values and coverage rates, so that missingness itself remains part of the operational signal.

\subsection{Data augmentation}

The number of naturally occurring model-window pairs is limited by the number of models and the observation period. To mitigate overfitting, we apply subsample augmentation: for each pair, we draw 80\% of its cases without replacement and recompute the signature, repeating $K$ times, which increases effective training size by $K{+}1$ while reflecting realistic sampling uncertainty \cite{shahzad2023estimation}.

\subsection{Multi-objective embedding learning}

The signature matrix is standardized and projected via PCA \cite{abdi2010principal} to a lower-dimensional intermediate space, which improves conditioning and provides a warm start for the learned linear projection; row-wise $L_2$ normalization then places all embeddings on the unit hypersphere. We return to the role of this warm start when interpreting our baseline comparison in Section~\ref{sec:forecasting}.

The projection is trained with three complementary objectives, combined into a single weighted loss. \textbf{Temporal contrastive learning} pulls a model's embedding at window $t$ toward its own embedding at window $t{+}1$, while pushing it away from a different-vendor model observed in the same window; restricting negatives to the same time window controls for platform-wide seasonal effects that would otherwise confound model-specific signals. \textbf{Cross-view reconstruction} splits the eight channels into a fault-side view (Channels 1, 2, 3, 6: what fails, when, how severely, and why) and a response-side view (Channels 4, 5, 7, 8: how quickly issues resolve, communication dynamics, routing, and workload context), and trains the embedding so that one view can be linearly reconstructed from the other; this ties the representation to both sides of the support lifecycle rather than only the more visible fault side. \textbf{Generational ordinality} is a weak supervision signal that only asks the embedding to preserve the relative ordering of successive versions within the same vendor (e.g., version $i$ closer to $j$ than to a later version $k$), without requiring precise inter-version distances, so that version evolution traces an interpretable trajectory. A controlled ablation isolating the contribution of each term is an open item we discuss in Section~\ref{sec:limitations}.

\subsection{Experimental setup}

The dataset comprises more than 33,000 production support cases spanning seven LLM families over 26 months, aggregated into per-model, per-month signatures. We compare OpEmbed against three non-learned reference points sharing the same underlying signals: a \emph{global average} baseline, a \emph{same-family average} baseline, and a \emph{Raw sig KNN} baseline that runs $K$-NN \cite{steinbach2009knn} directly on the standardized 75-dimensional signature without any learned projection. The first two test whether OpEmbed beats simple heuristics platform teams already use informally; the third tests whether it beats the raw features themselves, holding input information fixed.

\textbf{Task 1: few-shot operational forecasting.} Leave-one-model-out evaluation: a held-out model's metrics are predicted via distance-weighted $K$-NN among remaining models in embedding space, measured by MAE across six targets (bug rate, escalation rate, consultation rate, median initial response time, transfer bounce rate, communication intensity) and by $R^2$ from predicted-vs-actual plots.

\textbf{Task 2: cross-model fault transfer.} We predict each model-window's fault-type distribution via KNN from other models, scored by Jensen--Shannon Divergence (JSD) and by Top-$K$ hit rate (overlap between predicted Top-$K$ and actual Top-3 fault types).

\subsection{Industrial deployment and lessons learned}

OpEmbed is designed as an offline decision-support layer for LLM service reliability reviews rather than as an automated replacement for support or SRE judgment. In our industrial setting, structured support metadata are periodically aggregated into model-month signatures and mapped into an operational embedding space. The resulting neighborhoods are used to support four practitioner-facing workflows: early workload characterization for newly launched models, retrieval of troubleshooting playbooks from behaviorally similar models, triage of models whose support intensity or escalation profile is changing, and monitoring of embedding drift as a signal of structural shifts in service behavior.

Building the tool exposed several practical lessons. First, metadata-only signals are often sufficient for useful reliability profiling, which reduces privacy and confidentiality risk because case text and customer content are not required. Second, support traffic is not homogeneous: routine quota-related requests can dominate volume and suppress the variance of escalation-oriented outcomes, so separating technical from administrative workload is important. Third, the most useful representation is not necessarily the one with the cleanest visual clusters, but the one that improves concrete decisions such as staffing, onboarding readiness, and fault-pattern retrieval.

\section{Results}

\subsection{Operational fingerprints form structured, interpretable geometry}

\begin{figure}[htbp]
    \centering
    \includegraphics[width=0.48\textwidth]{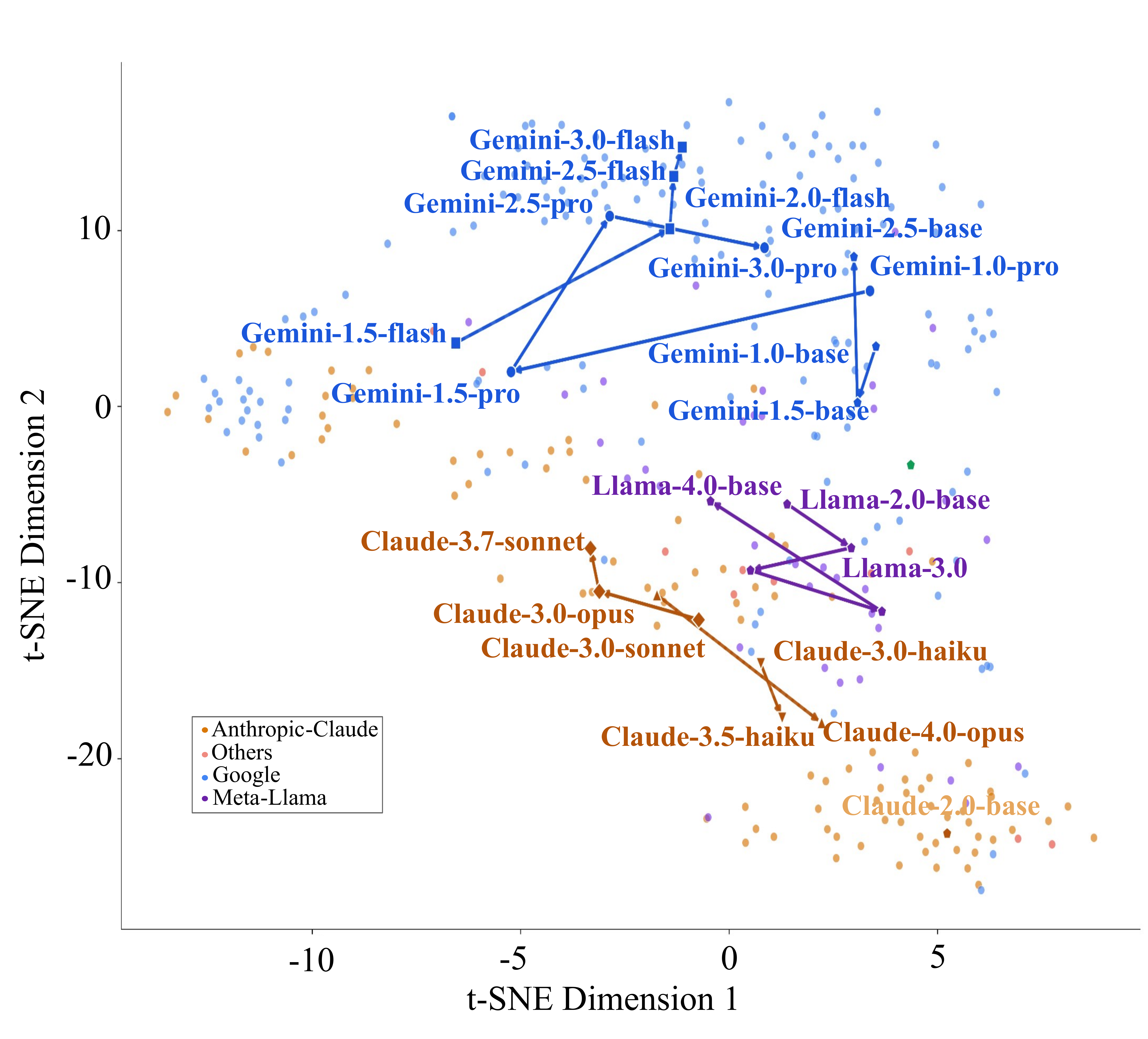}
    \caption{OpEmbed landscape with version evolution trajectories.}
    \label{fig2}
\end{figure}

As shown in Fig.~\ref{fig2}, OpEmbed forms a structured latent space rather than an arbitrary projection: Google models occupy a relatively concentrated region, Anthropic-Claude models form a denser cluster, and Meta-Llama models occupy a separate intermediate area, with several version lines tracing smooth local trajectories. This is consistent with the contrastive and ordinal objectives, which keep adjacent versions of a model line behaviorally related while allowing operational differences to emerge. We read this visualization (t-SNE, primarily qualitative) as evidence of organization and continuity, most convincing alongside the downstream results below.

\subsection{OpEmbed improves leave-one-model-out operational forecasting}
\label{sec:forecasting}

OpEmbed achieves lower forecasting error than all three non-learned baselines across all six targets. The comparison against Raw sig KNN is the most informative, since that baseline shares OpEmbed's exact inputs: any gap therefore reflects something added on top of the raw features. We note, however, that the projection is initialized from PCA on the standardized signature, and our current baselines do not include a PCA-projection-only KNN as an intermediate reference point. We can therefore attribute the gain over Raw sig KNN to the learned projection as a whole, but we cannot yet separate how much stems from the three training objectives versus the PCA-based dimensionality reduction it is initialized from; we treat isolating this via a PCA-only baseline as a concrete next step (Section~\ref{sec:limitations}).

As a robustness check, Table~\ref{tab:robustness} replicates the evaluation on a technical-support subset excluding quota-increase requests (5,534 cases, $\sim$16\% of volume). On the full dataset, Bug Rate, IRT Median, Bounce Mean, and Communication Intensity reach $R^2$ from 0.503 to 0.807, while Escalation and Consultation Rate are near zero because quota requests suppress their variance; filtering to technical cases reverses this (Escalation $R^2$: 0.031 $\to$ 0.592; Consultation: 0.064 $\to$ 0.483), while Bug Rate $R^2$ drops to 0.437. Different operational targets are thus predictable under different workload compositions, and OpEmbed remains informative across all six metrics with varying strength.

\begin{table}[htbp]
\centering
\caption{Prediction quality ($R^2$), full dataset vs. technical-only subset. Bold = better setting per metric.}
\label{tab:robustness}
\renewcommand{\arraystretch}{1.1}
\resizebox{\columnwidth}{!}{
\begin{tabular}{lccc}
\toprule
\textbf{Metric} & \textbf{Full $R^2$} & \textbf{Tech-only $R^2$} & \textbf{Best} \\
\midrule
Bug Rate         & \textbf{0.807} & 0.437 & Full \\
Escalation Rate  & 0.031 & \textbf{0.592} & Tech-only \\
Consult Rate     & 0.064 & \textbf{0.483} & Tech-only \\
IRT Median       & \textbf{0.595} & 0.331 & Full \\
Bounce Mean      & \textbf{0.657} & 0.011 & Full \\
Comm Intensity   & \textbf{0.503} & 0.072 & Full \\
\bottomrule
\end{tabular}
}
\end{table}

\subsection{Useful forecasts emerge from limited early operational history}

\begin{figure*}[htbp]
    \centering
    \includegraphics[width=0.85\textwidth]{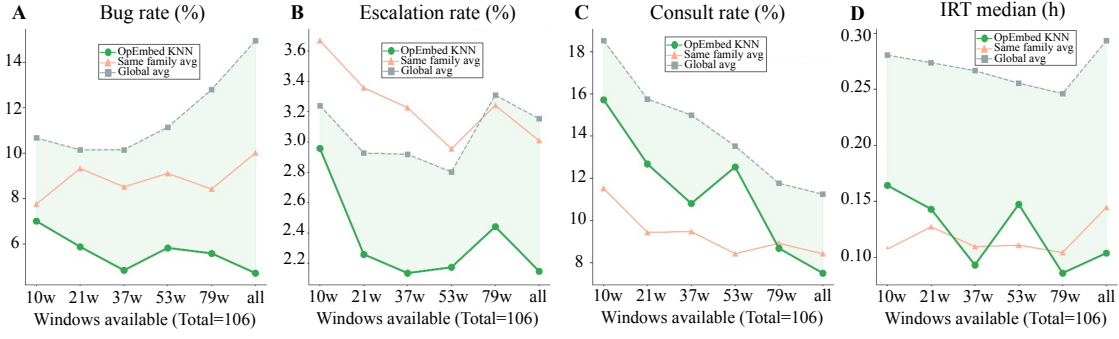}
    \caption{Data efficiency of operational forecasting with partial early histories, comparing OpEmbed KNN with same-family and global-average baselines across four targets.}
    \label{fig5}
\end{figure*}

Fig.~\ref{fig5} shows that OpEmbed already outperforms the global-average baseline at the earliest observation point across all four targets, i.e., a short prefix of incident history already places a new model into a meaningful operational neighborhood. For Bug Rate, MAE falls from about 7.0 at 10 windows to below 5.0 with all 106 windows; Escalation Rate, Consult Rate, and IRT Median show similar (though not perfectly monotonic) trends. This makes the framework directly relevant to early-stage deployment monitoring, when full operational histories are not yet available.

Here, ``windows'' refers to available model-month windows pooled across models, not calendar months for a single model.

\subsection{OpEmbed improves cross-model transfer of fault knowledge}

\begin{figure*}[htbp]
    \centering
    \includegraphics[width=0.78\textwidth]{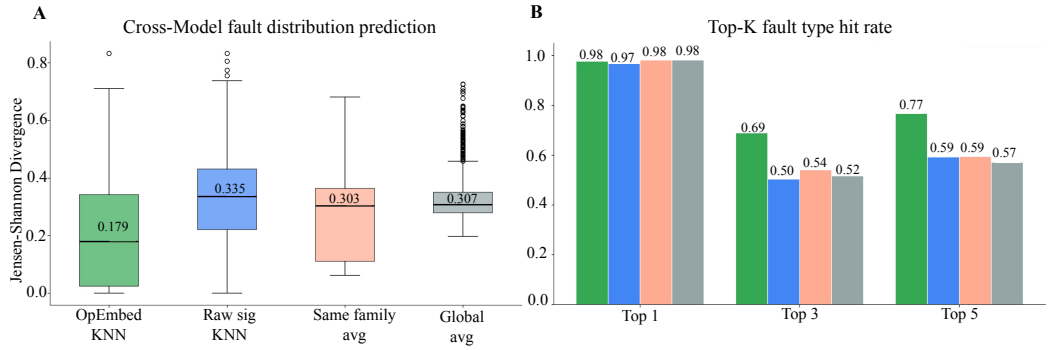}
    \caption{Cross-model fault transfer in distributional and Top-$K$ evaluation.}
    \label{fig6}
\end{figure*}

Fig.~\ref{fig6} evaluates cross-model transfer. In the distributional setting, OpEmbed achieves the lowest median JSD ($\approx$0.179) versus Raw sig KNN (0.335), Same-family average (0.303), and Global average (0.307), indicating it captures transferable structure beyond the single dominant fault type. At Top-1, all methods are near ceiling (0.97--0.98); the gap widens sharply at Top-3 and Top-5, where OpEmbed reaches 0.69 and 0.77 versus roughly 0.50--0.59 for the baselines --- more operationally meaningful than Top-1, since support teams benefit from a short ranked list rather than a single guess. Because OpEmbed clearly outperforms the same-family baseline, the result also rules out a trivial vendor-identity explanation: if OpEmbed merely encoded family labels, same-family averaging would have been hard to beat.

Table~\ref{tab:case_studies} shows five representative cases where OpEmbed's nearest-neighbor retrieval identifies all three actual fault types for the target model (3/3), while Raw sig KNN and Same-family baselines are less consistent. Notably, for deepseek-v3 the three nearest neighbors are all Claude models yet correctly predict the fault distribution, and for gemini-2.0-flash in 2026-01 OpEmbed anticipates a shift from the common Quotas category to 429 Errors through similarity to gemini-2.5-flash and gemini-3.0-flash, which had already exhibited this pattern --- a shift the global and same-family baselines fail to anticipate.

\begin{table*}[htbp]
    \centering
    \caption{Qualitative case studies of fault-spectrum transfer.}
    \label{tab:case_studies}
    \renewcommand{\arraystretch}{1.3}
    \resizebox{\textwidth}{!}{
    \begin{tabular}{ll p{4.3cm} p{4.6cm} ccc}
        \toprule
        \textbf{Target Model} &
        \textbf{Window} &
        \textbf{Actual Top-3 Faults} &
        \textbf{OpEmbed Neighbors Found} &
        \makecell{\textbf{OpEmbed}\\\textbf{Hit}} &
        \makecell{\textbf{Raw sig}\\\textbf{Hit}} &
        \makecell{\textbf{Same family}\\\textbf{Hit}} \\
        \midrule
        gemini-1.5-flash & 2024-05 & Quotas, Gemini, Model Behavior and Quality & claude-3.7-sonnet, gemini-1.5-pro & \textbf{3/3} & 1/3 & 2/3 \\
        claude-3.0-sonnet & 2024-10 & Quotas, Claude Models, Model Behavior and Quality & claude-3.5-sonnet, claude-3.7-sonnet & \textbf{3/3} & 2/3 & 2/3 \\
        llama-4.0 & 2025-04 & Quotas, Vertex AI Agent, Model Behavior and Quality & claude-3.0-haiku, claude-3.5-sonnet & \textbf{3/3} & \textbf{3/3} & 1/3 \\
        gemini-2.0-flash & 2026-01 & 429 Errors, Quotas, Vertex AI Model Registry & gemini-2.5-flash, gemini-3.0-flash & \textbf{3/3} & 1/3 & 2/3 \\
        deepseek-v3 & 2026-02 & Quotas, Vertex AI Agent, Model Behavior and Quality & claude-4.0-opus, claude-3.5-sonnet, claude-3.5-haiku & \textbf{3/3} & 1/3 & 1/3 \\
        \bottomrule
    \end{tabular}
    }
\end{table*}

\section{Conclusion}

We introduced OpEmbed, a framework for learning operational fingerprints of LLM cloud services from structured production incident metadata. Across embedding analysis, leave-one-model-out forecasting, early-history data-efficiency experiments, and cross-model fault transfer, operational structure is both learnable and practically useful: the learned space organizes models into interpretable neighborhoods, improves forecasting over raw-signature and heuristic baselines, remains informative under limited early history, and transfers fault knowledge across models and vendors. Capability benchmarks alone are insufficient for understanding production reliability behavior; structured support metadata provide a complementary, practically valuable evaluation axis.

Three lessons stand out from building this tool in production. First, operational metadata alone (without case text) already supports useful representation learning, which makes the approach easier to deploy under enterprise privacy constraints. Second, not all service outcomes are equally predictable under all workload compositions --- escalation and consultation rates become substantially more informative once routine quota-driven cases are separated from genuinely technical traffic. Third, the most useful embedding is not necessarily the one with the most visually separated clusters, but the one that supports concrete downstream decisions such as onboarding, triage, and fault-pattern retrieval. In practice, this makes OpEmbed a decision-support layer for staffing, onboarding, and triage in large-scale LLM service operations, rather than only an analysis artifact.

\section{Limitations}
\label{sec:limitations}

The data come from a single support ecosystem, so some patterns may reflect local workflows rather than universal operational behavior; ecosystems with a different ticket schema would likely require the eight channels in Section~\ref{sec:signature} to be adapted rather than reused as-is. The case distribution is imbalanced, with quota-related requests dominating the corpus and reducing variance in several targets, and sparse fields limit the contribution of some channels. Methodologically, we have not yet isolated the marginal contribution of each of the three training objectives via a controlled ablation; the reported results reflect the combined objective used in our main evaluation, and decomposing its components is a concrete direction for follow-up work. Relatedly, because our projection is initialized from PCA, our comparison against Raw sig KNN does not by itself distinguish the contribution of the three learned objectives from that of the PCA-based dimensionality reduction; adding a PCA-projection-only KNN baseline is a straightforward extension we plan to include in future revisions. Finally, OpEmbed captures operational similarity rather than intrinsic model capability, and should complement, not replace, benchmark-based evaluation.

\section*{Acknowledgment}

The authors thank the Google Cloud Platform Customer Support data team for providing the secure operational data environment that made this research possible. This study was carried out under applicable internal data security and confidentiality requirements, with data handling restricted to case-level metadata only.

\section*{Data Availability}

The data used in this study consist of internal production support metadata derived from customer-facing cloud service operations. Access is restricted under Google's internal data governance and security review requirements as well as customer confidentiality obligations, and the raw data cannot be made publicly available. 

\bibliographystyle{IEEEtran}
\bibliography{sample-base}

\appendices 

\section{Supply information}
\label{app:information}

\begin{figure*}[!t]
    \centering
    \includegraphics[width=0.95\textwidth]{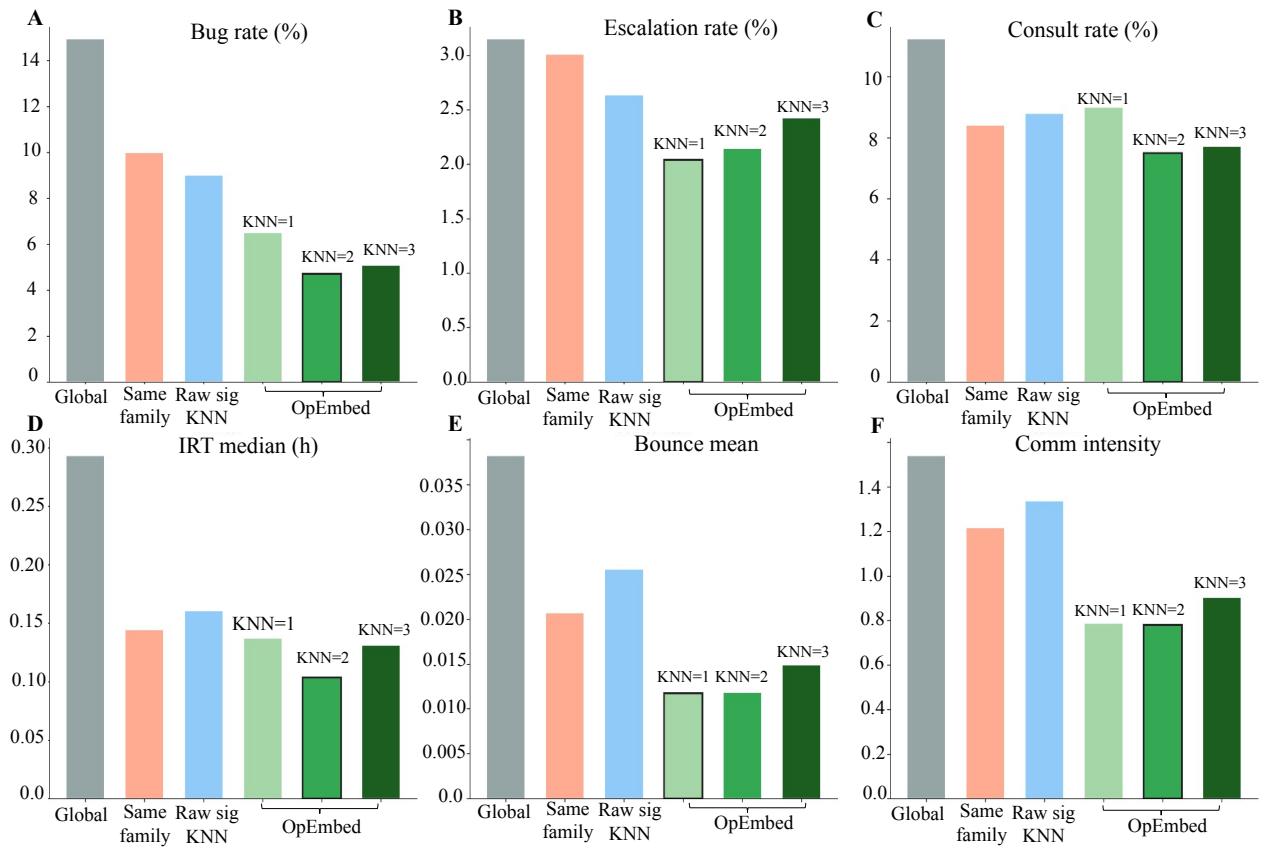}
    \caption{Leave-One-Model-Out few-shot forecasting across six operational targets}
    \label{app1}
\end{figure*}

\begin{figure*}[!t]
    \centering
    \includegraphics[width=0.95\textwidth]{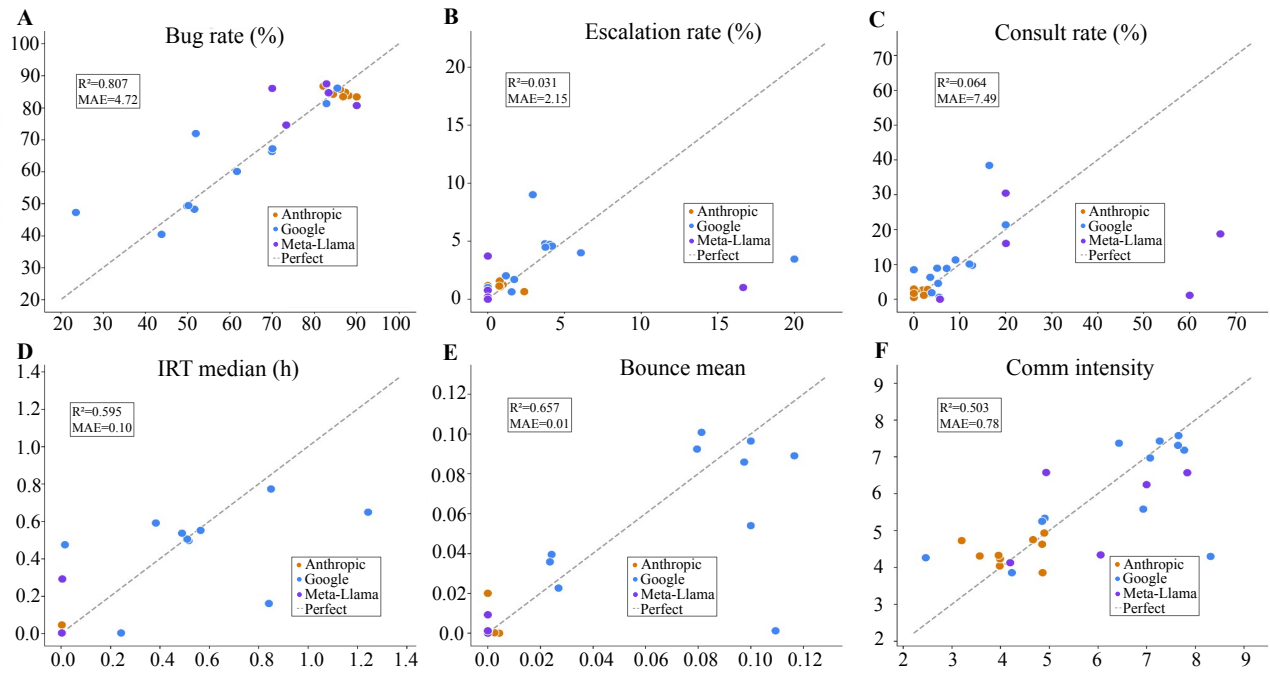}
    \caption{Predicted versus actual operational outcomes under leave-one-model-out forecasting}
    \label{app2}
\end{figure*}

Figures \ref{app1} and \ref{app2} show that OpEmbed consistently improves model-level operational forecasting. Across all six targets, OpEmbed achieves lower error than non-learned baselines, indicating that nearby points in the learned space tend to share similar downstream operational behavior. This advantage is most meaningful relative to Raw sig KNN, since the raw baseline already uses the same hand-engineered inputs: any gap between the two therefore reflects something added on top of the raw features, rather than access to richer information. We note, however, that our projection is initialized from PCA on the standardized signature, and our current baselines do not include a PCA-projection-only KNN as an intermediate reference point. As a result, we can attribute the gain over Raw sig KNN to ``the learned projection as a whole,'' but we cannot yet separate how much of that gain is due to the three training objectives versus the PCA-based dimensionality reduction the projection is initialized from. 

Table \ref{tab:nearest_neighbors} offers a profiling-oriented view of the learned space. Many nearest-neighbor relations are intuitively coherent. Within Anthropic, pairs such as claude-3.5-sonnet / claude-3.7-sonnet (0.044) and claude-3.0-haiku / claude-3.0-sonnet (0.084) indicate that OpEmbed preserves fine-grained operational similarity among closely related variants. A similar pattern appears within Gemini, including gemini-2.0-flash / gemini-2.5-pro (0.046), gemini-2.5-flash / gemini-2.5-pro (0.071), and gemini-1.5-flash / gemini-1.5-pro (0.077).

At the same time, Table \ref{tab:nearest_neighbors} also contains informative cross-family neighbors. For example, deepseek-v3 is closest to claude-3.0-sonnet, and llama-3.3 is closest to claude models, rather than to another Llama model. These cases matter because they show that operational similarity does not always follow vendor lineage. Instead, models from different families may converge in the learned space when their observed support behavior is more alike.

\begin{table*}[!t]
    \centering
    \caption{\textbf{Nearest neighbors in OpEmbed space} }
    \label{tab:nearest_neighbors}
    \renewcommand{\arraystretch}{1.2}
    
    \resizebox{\textwidth}{!}{
    \begin{tabular}{llcccc}
        \toprule
        \textbf{Model} & 
        \textbf{Family} & 
        \makecell{\textbf{Nearest Neighbor} \\ \textbf{(cosine dist)}} & 
        \makecell{\textbf{2nd Nearest} \\ \textbf{(cosine dist)}} & 
        \makecell{\textbf{3rd Nearest} \\ \textbf{(cosine dist)}} & 
        \makecell{\textbf{NN is} \\ \textbf{Same Fam?}} \\
        \midrule
        claude-2.0-base & Anthropic-Claude & claude-3.5-haiku (0.199) & claude-4.0-sonnet (0.257) & claude-3.0-haiku (0.317) & \ding{51} \\
        claude-3.0-haiku & Anthropic-Claude & claude-3.0-sonnet (0.084) & claude-3.5-haiku (0.114) & claude-3.5-sonnet (0.248) & \ding{51} \\
        claude-3.0-opus & Anthropic-Claude & claude-3.0-sonnet (0.241) & claude-3.0-haiku (0.292) & claude-3.5-sonnet (0.327) & \ding{51} \\
        claude-3.0-sonnet & Anthropic-Claude & claude-3.0-haiku (0.084) & deepseek-v3 (0.233) & claude-3.0-opus (0.241) & \ding{51} \\
        claude-3.5-haiku & Anthropic-Claude & claude-4.0-opus (0.109) & claude-3.0-haiku (0.114) & claude-2.0-base (0.199) & \ding{51} \\
        claude-3.5-sonnet & Anthropic-Claude & claude-3.7-sonnet (0.044) & claude-3.0-sonnet (0.242) & claude-3.0-haiku (0.248) & \ding{51} \\
        claude-3.7-sonnet & Anthropic-Claude & claude-3.5-sonnet (0.044) & claude-3.0-sonnet (0.247) & claude-3.0-haiku (0.337) & \ding{51} \\
        claude-4.0-opus & Anthropic-Claude & claude-3.5-haiku (0.109) & claude-4.0-sonnet (0.235) & claude-3.0-haiku (0.275) & \ding{51} \\
        claude-4.0-sonnet & Anthropic-Claude & claude-4.0-opus (0.235) & claude-2.0-base (0.257) & claude-3.5-haiku (0.292) & \ding{51} \\
        deepseek-v3 & Deepseek & claude-3.0-sonnet (0.233) & llama-3.3 (0.372) & claude-3.0-haiku (0.395) & \ding{55} \\
        gemini-1.0-base & Google & gemini-1.0-pro (0.308) & gemini-1.5-base (0.470) & gemini-2.0-flash (0.472) & \ding{51} \\
        gemini-1.0-pro & Google & gemini-1.0-base (0.308) & gemini-2.5-base (0.324) & gemini-1.5-base (0.346) & \ding{51} \\
        gemini-1.5-base & Google & gemini-1.0-pro (0.346) & gemini-1.0-base (0.470) & llama-3.0 (0.523) & \ding{51} \\
        gemini-1.5-flash & Google & gemini-1.5-pro (0.077) & gemini-2.5-pro (0.230) & gemini-2.0-flash (0.389) & \ding{51} \\
        gemini-1.5-pro & Google & gemini-1.5-flash (0.077) & gemini-2.5-pro (0.379) & gemini-2.0-flash (0.477) & \ding{51} \\
        gemini-2.0-flash & Google & gemini-2.5-pro (0.046) & gemini-2.5-flash (0.061) & gemini-3.0-flash (0.131) & \ding{51} \\
        gemini-2.5-base & Google & gemini-3.0-pro (0.250) & gemini-3.0-flash (0.295) & gemini-1.0-pro (0.324) & \ding{51} \\
        gemini-2.5-flash & Google & gemini-2.0-flash (0.061) & gemini-2.5-pro (0.071) & gemini-3.0-flash (0.120) & \ding{51} \\
        gemini-2.5-pro & Google & gemini-2.0-flash (0.046) & gemini-2.5-flash (0.071) & gemini-3.0-flash (0.140) & \ding{51} \\
        gemini-3.0-flash & Google & gemini-2.5-flash (0.120) & gemini-3.0-pro (0.127) & gemini-2.0-flash (0.131) & \ding{51} \\
        gemini-3.0-pro & Google & gemini-3.0-flash (0.127) & gemini-2.5-flash (0.179) & gemini-2.0-flash (0.238) & \ding{51} \\
        gemma-2 & Google & deepseek-v3 (0.410) & llama-3.3 (0.410) & claude-2.0-base (0.579) & \ding{51} \\
        llama-2.0-base & Meta-Llama & llama-3.0 (0.228) & deepseek-v3 (0.459) & llama-4.0 (0.475) & \ding{51} \\
        llama-3.0 & Meta-Llama & llama-2.0-base (0.228) & llama-3.3 (0.477) & gemini-1.5-base (0.523) & \ding{51} \\
        llama-3.2 & Meta-Llama & llama-4.0 (0.163) & claude-4.0-opus (0.289) & llama-3.3 (0.304) & \ding{51} \\
        llama-3.3 & Meta-Llama & claude-4.0-sonnet (0.311) & claude-4.0-opus (0.317) & deepseek-v3 (0.372) & \ding{55} \\
        llama-4.0 & Meta-Llama & llama-3.2 (0.163) & deepseek-v3 (0.429) & llama-2.0-base (0.475) & \ding{51} \\
        
        \bottomrule
    \end{tabular}
    }
\end{table*}

\begin{figure*}[!t]
    \centering
    \includegraphics[width=0.95\textwidth]{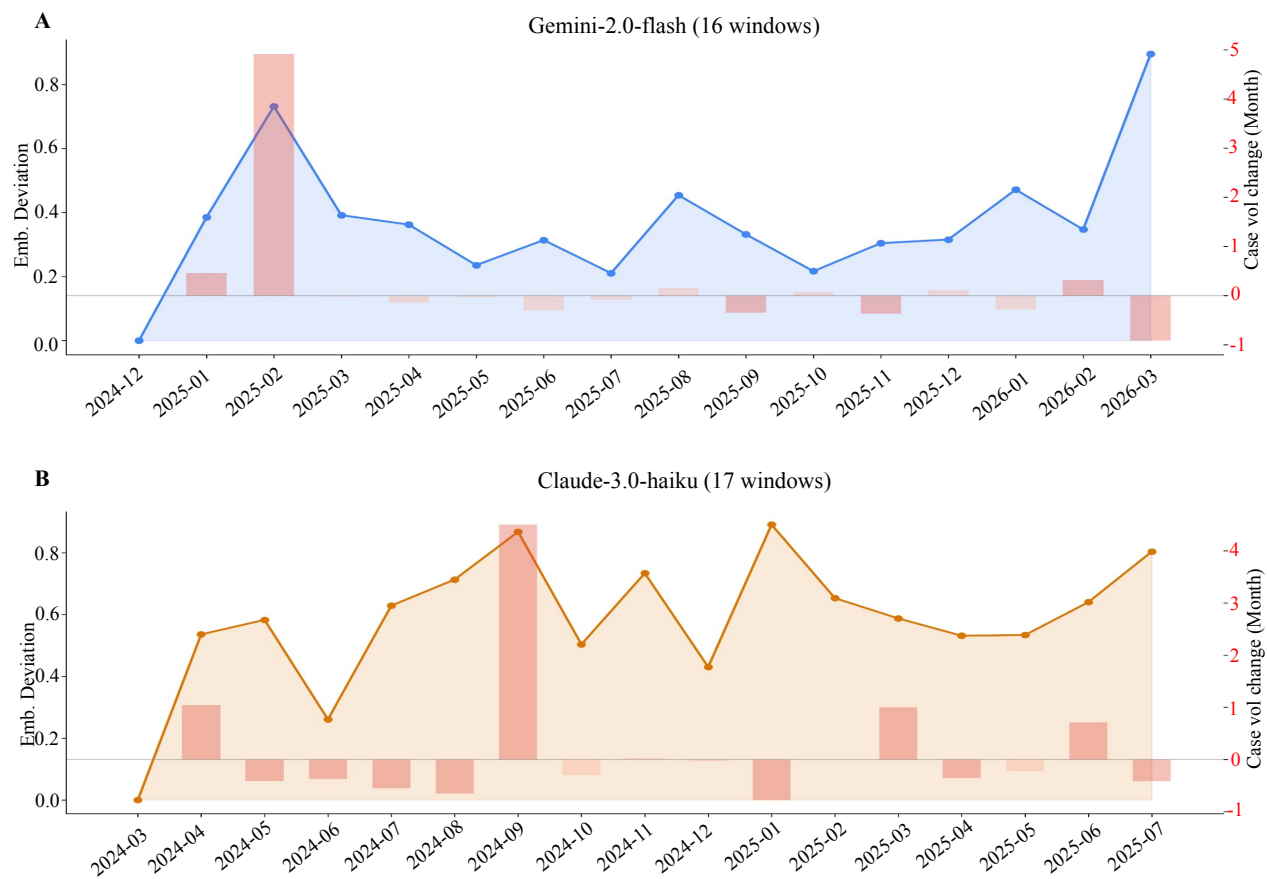}
    \caption{Representative temporal drift diagnostics for two models. Embedding deviation is shown together with month-over-month case-volume change as contextual workload information only. The figure is illustrative rather than predictive: deviation indicates operational change, not necessarily degradation, and case-volume change is not a forecast target.}
    \label{fig7}
\end{figure*}

Figure \ref{fig7} provides two representative monitoring examples of online embedding deviation over time. In both cases, the deviation trajectory highlights windows in which the model's operational behavior moves away from its recent history, suggesting a structural change rather than ordinary month-to-month fluctuation. Because the deviation is computed relative to the preceding windows, the view is naturally suited to online monitoring settings in which future data are unavailable.

The purpose of this analysis is descriptive rather than predictive. A high deviation does not necessarily indicate worse service quality; it may instead reflect model updates, workload redistribution, customer-mix changes, or emerging support patterns. For this reason, Figure \ref{fig7} should be interpreted as an operational monitoring view that helps platform teams identify windows worth inspection, rather than as a formal validation of degradation. We show two representative cases for readability, while the full set of monitored models exhibits the same intended usage pattern: highlighting behavior shifts that may warrant further operational review. Taken together, these results suggest that the framework is scalable to heterogeneous model portfolios and can generalize to other support ecosystems that expose comparable structured incident metadata, even when case text cannot be used.

\section{Acceptance Notification}
\label{app:acceptance}

\begin{figure*}[!t]
    \centering
    \includegraphics[width=0.95\textwidth]{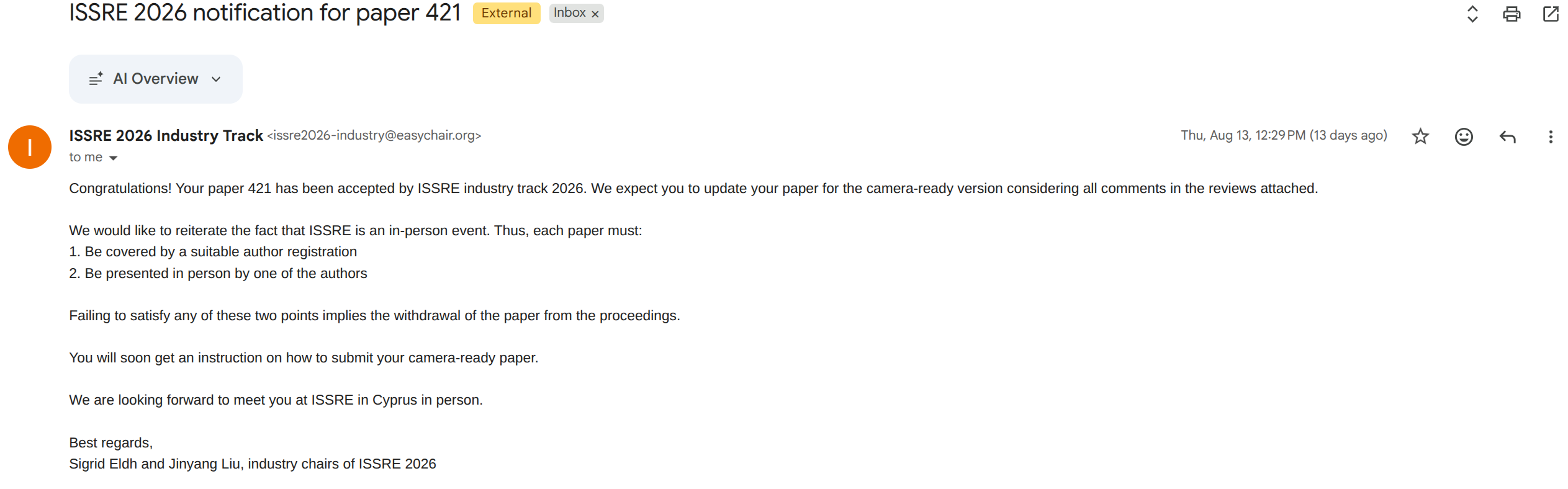}
    \caption{ISSRE 2026 notification for paper 421}
    \label{app3}
\end{figure*}

\section{Peer Reviews}
\label{app:reviews}

Below we provide the full, unedited peer reviews received during the ISSRE 2026 review process to certify the peer-reviewed status of this manuscript.

\subsection*{Reviewer 1}
\noindent\textbf{Overall Recommendation:} 3 (Weak Accept) \\
\textbf{No use of AI:} Yes

\vspace{0.5em}
\noindent\textbf{Summary:} \\
The paper proposes OpEmbed, a way to learn a compact operational fingerprint of an LLM cloud service purely from structured support case metadata, no case text. For each model and month it builds a 75 dimensional signature across eight channels, and learns a low dimensional embedding using contrastive learning, cross view reconstruction, and a version ordering constraint. It's evaluated on 33,000+ Google Cloud support cases across seven model families over 26 months, and shows the embedding organizes models sensibly, forecasts operational metrics for a held out model and transfers fault patterns across models.

\vspace{0.5em}
\noindent\textbf{Strengths:} \\
The core insight is genuinely nice ... capability benchmarks say nothing about how much support load a model will generate once it's live, and this reframes that operational behavior as something you can learn from data you already collect. The privacy angle is a real strength too, since using only metadata and never case text makes it far easier to deploy under enterprise constraints. The evaluation is honest and does the right comparison. And the paper is candid about its own confounds, especially the PCA warm start question.

\vspace{0.5em}
\noindent\textbf{Shortcomings:} \\
The paper is honest about its gaps but that doesn't close them. The biggest is isolating the three training objectives is explicitly deferred to future work, so we don't actually know which parts of the fairly complex loss are doing the work ... for a method paper that's the central question. The PCA confound compounds this: they admit the gain over Raw sig KNN could partly come from the PCA step rather than the learned objectives, and the PCA only baseline that would settle it is missing. The results are also uneven in a way the averages hide with escalation and consultation R2 near zero on the full data and only recovering after filtering to technical cases. And it's one vendor's support ecosystem over one 26 month window, so how much transfers elsewhere is unclear. The interpretability claims lean on a qualitative t-SNE plot, which is suggestive but not evidence.

\vspace{0.5em}
\noindent\textbf{Overall:} \\
The PCA only baseline is one of the main thing missing.

\vspace{1em}
\subsection*{Reviewer 2}
\noindent\textbf{Overall Recommendation:} 1 (Reject) \\
\textbf{No use of AI:} Yes

\vspace{0.5em}
\noindent\textbf{Summary:} \\
The paper presents OpEmbed, an approach that represents LLMs with signature (embeddings) constructed from 75 features over model month periods. The evaluation uses Google Cloud support cases covering seven LLM families and 26 months. The signature representations are used for neighbour-based prediction of operational metrics and cross-model transfer of fault distributions. However, the current evaluation does not establish that OpEmbed provides leakage-free forecasting or sufficient practical benefit over simpler approaches. The presentation of material is confusing.

\vspace{0.5em}
\noindent\textbf{Strengths:} 
\begin{itemize}
    \item Motivation of the paper is well presented in the Introduction.
    \item Related work section is adequate.
    \item It has good industry and reliability relevance.
    \item The omission of case text is attractive from privacy perspective.
\end{itemize}

\vspace{0.5em}
\noindent\textbf{Weaknesses:} \\
The target leakage can be a possible major issue with this approach. The signature includes fault spectrum, severity, resolution efficiency, interaction dynamics, root cause, etc. The predicted targets include bug rate, escalation rate, consultation rate and other which appear to correspond directly (e.g. fault spectrum and bug rate) to the input channels. The paper does not explicitly state that target features are removed or that only earlier time windows are used to predict later outcomes.

The practical question is whether OpEmbed is preferable to simply retrieving previous close versions of LLMs. The authors try to address this to an extent by introducing 'same-family average'. However, looking at their Fig.2 family-wise clustering can be quite coarse, whereas models of similar generations (e.g. Claude 3, Gemini 2.5) are much closer. This raises a question if finer grouping is a more competitive comparison.

The authors refer to 7 LLM families, yet these are never properly defined leaving the reader guessing. For example, in Fig.2 one could count 3 families (Gemini, Llama, Claude) in Table 2 - 4 families (Gemini, Claude, DeepSeek, Llama)...

The results from Figure 3 are overstated. While the approach seems better for bug rate and marginally better for escalation rate, it is clearly worse for consult rate as compared to 'same-family' (which is the best) over 79 weeks, and is comparable /mixed to 'same-family' for IRT median.

In Fig 4, Top-K is never formally defined, making meaningful interpretation of the results difficult.

Lessons learned appear too vague and not tied explicitly to empirical data.

Research questions are not properly defined and instead unconventional 'task1' and 'task2' are used.

Calling $R^2$ 'prediction quality' seems very informal and misleading, when conventionally it is a proportion of variance in predicted variable as explained by predictor variable.

\vspace{0.5em}
\noindent\textbf{Industry relevance:} Good fit. \\
\textbf{Significance / novelty:} There is a potential novelty in this approach, but it has to be convincingly demonstrated. Also, it would be interesting to see any operational impact that this approach might have. Currently this is not presented. \\
\textbf{Soundness:} See weaknesses. \\
\textbf{Evaluation / verification:} No replication data is provided. 

\vspace{0.5em}
\noindent\textbf{Presentation:} \\
The paper is difficult to follow. It has unconventional structure, misses important details/explanation, is confusing, and has redundant parts. 
For example:
\begin{itemize}
    \item It is a good idea to separate 'Method' and 'Experimental Methodology' which are currently both in Section 3. Also, it is very surprising to see 'lessons learned' in section 3.6, which is part of 'Method'.
    \item Many terms and metrics are introduced without definitions leaving the reader guessing what these are: 'same-family average', 'top-k', 'raw sig', 'channels', 't-sne', etc.
    \item Figure 1 has very small font size and is hard to read.
    \item Figure 3 has confusing titles and the reader has to guess what is 'bug rate', for example. Is that proportion of bugs across all issues, proportion of resolved bugs, detected bugs, predicted bugs, or MAE of bug rate prediction.
    \item The last paragraph in 3.6 and the last paragraph in conclusion essentially repeat each other.
\end{itemize}

\vspace{0.5em}
\noindent\textbf{Related work and references:} Well presented.

\vspace{1em}
\subsection*{Reviewer 3}
\noindent\textbf{Overall Recommendation:} 4 (Accept) \\
\textbf{No use of AI:} Yes

\vspace{0.5em}
\noindent\textbf{Detailed Comments to the Authors:} \\
This paper fits ISSRE's scope well; it addresses the "ilities" of services underlying LLM models that benchmarks and leaderboards don't capture, since they focus solely on capabilities. I found this paper intriguing in the sense that the authors have started addressing what it means to operate AI services reliably at scale in the context of models that need to be updated and onboarded continuously.

The key idea put forward in this paper is an "operational fingerprint" (OpEmbed) learned from support metadata that can forecast future support burden and transfer fault knowledge across models and vendors. The authors support their claims with results from real production data. However, although they claim that OpEmbed could help support decisions, they only cover these claims hypothetically and qualitatively. There are no real examples of onboarding or staffing decisions made using OpEmbed; no description of a threshold or workflow (e.g., "if operational distance to nearest neighbor exceeds X, trigger Y"), or outcome data (e.g., "using this playbook-retrieval workflow reduced resolution time by Z").

\vspace{0.5em}
\noindent\textbf{Positive:}
\begin{itemize}
    \item The cross-model, cross-vendor "operational fingerprint" derived from structured, non-text metadata is a solid contribution distinct from case-level AIOps work. The privacy-by-design approach is an important enabler of adoption.
    \item Substantial, realistic industrial evaluation base (33,000+ real production support cases across seven LLM families over 26 months). Most academic AIOps papers would have a hard time matching this scale and realism.
    \item The paper openly addresses its limitations—specifically, by flagging the missing PCA-only ablation baseline and the unisolated contributions of its three training objectives.
\end{itemize}

\vspace{0.5em}
\noindent\textbf{Negative:}
\begin{itemize}
    \item The main weakness is the leap from "useful signal" to "decision support layer," which is left unsubstantiated. As noted above, there are no results from real decisions or gains from applying OpEmbed in staffing, onboarding, and triage. The results cover only the analysis.
    \item I had to deduce some acronyms from context because they were not defined on first use, for example, PCA, KNN, MAE, and IRT. That shouldn't be imposed on readers.
    \item The realism of the data used for evaluation brings reproducibility challenges. This is more of an open question for the AIOps community; nonetheless, it would be nice to see the authors' thoughts on this respect.
\end{itemize}

\end{document}